# Identifications of concealed weapon in a Human Body

Prof. Samir K. Bandyopadhyay[1], Biswajita Datta[2], Sudipta Roy[3]

[1,3] Department of Computer Science and Engineer, University of Calcutta,
92 A.P.C. Road, Kolkata-700009, India.
[2] Dept. of Computer Science & Engineering,
St. Thomas College of Engineering & Technology, Kolkata, India
skb1@vsnl.com[1], biswa.jita@gmail.com[2], sudiptaroy01@yahoo.com[3],

**Abstract.** The detection of weapons concealed underneath a person's cloths is very much important to the improvement of the security of the public as well as the safety of public assets like airports, buildings, and railway stations etc. Manual screening procedure gives unsatisfactory results when the object is not in the range of security personnel and when there is an uncontrolled flow of people. The goal is to develop an automatic detection and recognition system of concealed weapons using sensor technologies and image processing. The focus of this paper is to develop a new algorithm using a colour visual image and a corresponding IR image for such a concealed weapon detection application by the help of fusion technology.
**Keywords:** concealed weapon detection, color image, IR image, DWT Image fusion.

**Introduction.**

A weapon is any object that can do harm to another individual or group of individuals. This definition not only includes objects typically thought of as weapons, such as knives and firearms, but also explosives, chemicals, etc. so this harmful things need to be detect for securing general public as well as public assets like airports and buildings etc. Already used manual screening procedure sometimes gives wrong alarm indication, and fails when the object is not in the range of security personnel as well as when it is impossible to manage the flow of people through a controlled procedure. It also disappoints us when we try to identify a person who is the victim of an accident in future. We have recently witnessed the series of bomb blasts in Mumbai, Delhi, and Guhahati etc. Bombs went off in buses and underground stations. And killed many and left many injured and left the world in shell shock and the Indians in terror. This situation is not limited to India but it can happen or already happened anywhere and anytime in the world. People think bomb blasts can't be predicted before handled. In all of these cases CWD by scanning the images gives satisfactory results. But no single sensor technology can provide acceptable performance. So we try to bring the eventual deployment of automatic detection and recognition of concealed weapons. It is a technological challenge that requires innovative solutions in sensor technologies and image processing. The problem also presents challenges in the legal arena; a number of sensors based on different phenomenology as well as image processing support are being developed to observe objects underneath people's clothing. Now image fusion has been identified as a key technology to achieve improved CWD procedures. In our current work we focus on fusing visual and low cost IR images for CWD.

Infrared images are depends on the temperature distribution information of the target to form an image. Usually the theory follows here is that the infrared radiation emitted by the human body is absorbed by clothing and then re-emitted by it. In the IR image the background is almost black with little detail because of the high thermal emissivity of body. The weapon is darker than the surrounding body due to a temperature difference between it and the body (it is colder than human body). The visual image is a mental image that is similar to a visual perception. The resolution in the visual image is much higher than that of the IR image. It is nothing but a RGB image that supports human visual perception. But there is no useful information on the concealed weapon in the visual image. The human visual system is very sensitive to colours. To utilize this ability if we apply this image with other image in fusion technique we get a better fused image that helps for detection.

**Brief Review**

Imaging techniques based on a combination of sensor technologies and processing will potentially play a key role in addressing the concealed weapon detection problem. One critical issue is the challenge of performing detection at a distance with high probability of detection and low probability of false alarm. Yet another difficulty to be surmounted is forging portable multisensory instruments. Also, detection systems go hand in hand with subsequent response by the operator, and system development should take into account the overall context of deployment [1]. Concealed Weapon using the radar image are proposed by Yu-Wen Chang et all [2,3] in which drawbacks such as glint and specular reflection or artifacts such as coherent interference these problems should be able to be overcome. A new algorithms proposed by Zhiyun Xue et all[6] in which fuse a color visual image and a corresponding IR image for such a concealed weapon detection application in which they have great success. So fusion is an important step, we use here DWT fusion , some more improve method are there such as Chu-Hui Lee et all[13] produce a easy applications to adjust for anytime, and anywhere you like, make sure that may work and take a photograph nicely. The DWT fusion methods provide computationally ancient image fusion techniques Various fusion rules for the selection and combination of sub band coefficients increase the quality perceptual and quantitatively measurable of image fusion in specific applications. For binaries the fused image there are several method[8-10] Otsu method are chosen because this method are global method and effective for this type of image. The concept of small area removal are taken from[4]. However, based on biological research results, the human visual system is very sensitive to colours. To utilize this ability, some researchers map three individual monochrome multispectral images to the respective channels of an RGB image to produce a false color fused image. In many cases, this technique is applied in combination with another image fusion procedure. Such a technique is sometimes called color composite fusion. we present a new technique to fuse a color visual image with a corresponding IR image for a CWD application.

**Proposed Method**

In our proposed technique for CWD we consider two types of image – a visual image and an IR image. Visual image is nothing but an RGB image which has three main colour components Red, Green and Blue. Since the human visual system is very sensitive to colours this image creates a natural perception of an object to human vision but not helps so much in the detection of concealed weapon. For this we consider IR image as second input. It basically depends on high thermal emissivity of the body. Basically the infrared radiation emitted by the body is absorbed by clothing and then re-emitted by it, is sensed by the infrared sensors. Due to difference in thermal emissivity we can realize the hidden object but since the background is almost black this image cannot help in CWD alone.

*Resize two input images:* Since these two input images are taken from two different image sensing devices so they are of different size. So we first resize these two types of images because the image fusion and other operations are not possible if the sizes are not same.

*Combine two images:* Perform the addition operation between visual and IR (visual + IR) images to get the $I_{v\_IR}$ image. But the resultant image does not give enough information. Then we complement the IR image ($I_{IR\_c}$) to remove the background darkness. IR image lies the intensity between 0 to 255 intensity thus complement means subtracting all matrix component from 255 and we get complemented form or reverse form of the IR image. Then add visual image and complemented IR image (visual + complemented IR) and get a resultant image which is denoted by $I_{v\_IR\_c}$.

*Conversion of IR to HSV:* Then we convert IR image into HSV colour model ($I_{IR\_HSV}$) because components of IR image are all correlated with the amount of light hitting the object, and therefore with each other, image descriptions in terms of those components make object discrimination difficult. Descriptions in terms of hue/lightness/saturation are often more relevant.

*Fused two images:* After converting HSV model the image is now three components. Now we can use fusion technique because two images have the same dimension with same size and we use DWT fusion technique between HSV colour image ($I_{IR\_HSV}$) and combined image $I_{v\_IR\_c}$.

Processing for showing the weapon clearly in the visual image: Then this fused image converted into gray scale image. Now we use Otsu's local thresholding technique for binarizing fused gray scale image.

Then Extract the weapon portion by calculating all connected area component and remove too small component and also too large component according to the area values.

To show the weapon in the actual RGB visual image we multiply the weapon's binary images with three dimensional RGB image. Basically the element wise multiplication is performed between two matrices.

Now contour detection is used to detect edges of weapon from the weapon binary image and we use canny edge detector for detecting the edges.

Then this binarizes contour image is divided into three components and multiply as before and we get contour with visual RGB image where we can detect the concealed weapon under the person's clothes very clearly.

Here below is the flow diagram of our proposed method is shown.

**Algorithm:**
Step 1: Take a visual image (basically, RGB image) and an infrared (IR) image as input.
Step 2: Resize this two image so that they have same size.
Step 3: Combine i.e. add resized Visual and IR image.
Step 4: Complement the IR image.
Step 5: Combine i.e. add resized Visual image and complemented IR image.
Step 6: Convert the visual RGB image to its HSV format.
Step 7: Perform DWT fusion on Step 5's combined image and Step 6's converted HSV image.
Step 8: Convert the fused image into its gray scale format.
Step 9: Binarize the Fused image.
Step 10: Detect the weapon from that image.

Step 11: Combine this detected weapon with visual image.
Step 12: For detecting the weapon clearly we find out the contour of the weapon.
Step 13: Then combine the contour of the weapon with visual image.
Step 14: End

**Result & Analysis**

The weapons detection algorithm consists of several steps which will be explained in detail in the following. Take two images in the same pose visual RGB image and IR image which are shown in **figure 1** and **figure 2**. Resize these two types of image because image fusion and addition are not able to perform if the sizes are not same..

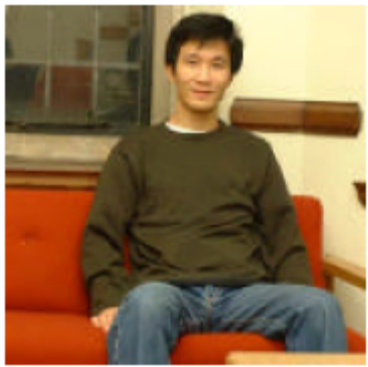
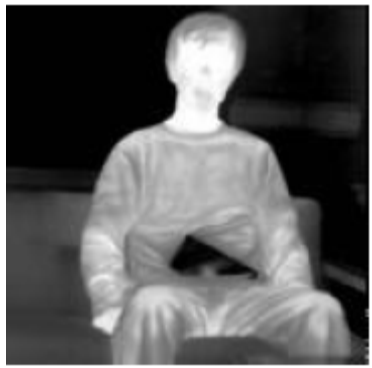
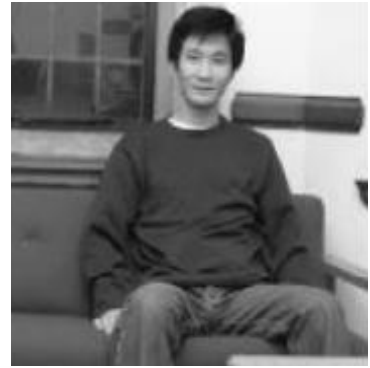

Figure 1 : RGB image        Figure 2 : IR image        Figure 3 : Gray image

Though this part of the gray scale conversion are not take in to consideration in our algorithms to find out concealed weapon. This part is to visual comparison between one dimension IR image and gray image.

Combine basically add visual image and IR image and the result is shown in **figure 4**. Actually we want to detect the hiding details from **figure 4** but image from **figure 4** is hazy, so we do not get enough information from **figure4**. Complement the IR image which is use full in the next operation and this complement image is shown in **figure 5.** IR image lies the intensity between 0 to 255 intensity thus complement means subtracting all matrix component from 255 and we get complemented form or reverse form of the IR image. Then add visual image and complemented IR image which is shown in **figure 6**.

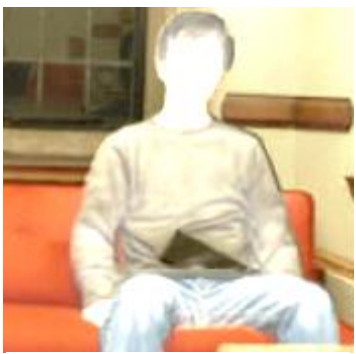
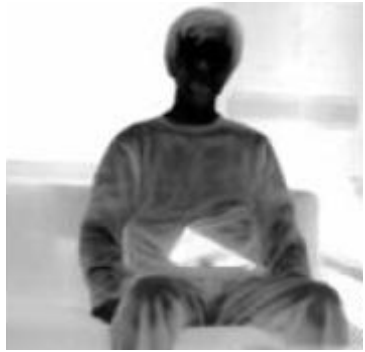
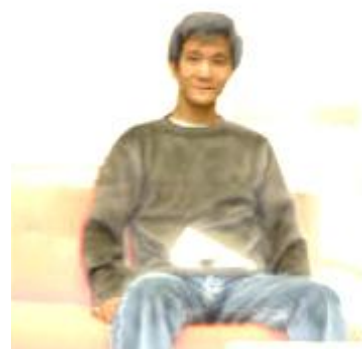

Figure 4: Combined image        Figure 5: Complemented IR        Figure 6: Combined1 image

In this steps fusion is not possible due to dimension mismatch. We do these steps because in this step difference between hiding details and man are recognizable. Then we convert IR image into HSV colour model and it is shown in **figure7** because components of IR image are all correlated with the amount of light hitting the object, and therefore with each other, image descriptions in terms of those components make object discrimination difficult. Descriptions in terms of hue/lightness/saturation are often more relevant. After converting HSV model the image is now three components. Now we can use fusion technique because two images have the same dimension with same size. Then we use DWT fusion technique between HSV color image and combined image is shown in **figure 8.** The discrete wavelet transform DWT is a spatial frequency decomposition that provides a flexible multi resolution analysis of an image. In wavelet transformation due to sampling, the image size is halved in both spatial directions at each level of decomposition process thus leading to a multi1resolution signal representation. The advantages of image fusion over visual comparison of multi-modality are: (a) the fusion technique is useful to correct for variability in orientation, position and dimension; (b) it allows precise

anatomic1physiologic correlation; and (c) it permits regional quantisation. Many image processing like de-noising, contrast enhancement, edge detection, segmentation, texture analysis and compression can be easily and successfully performed in the wavelet domain. Wavelet techniques thus provide a powerful set of tools for image enhancement and analysis together with a common framework for various fusion tasks. Applying fusion technique image sharpness and contrast enhanced. Then this fused image converted into gray scale image is shown in **figure 9**.

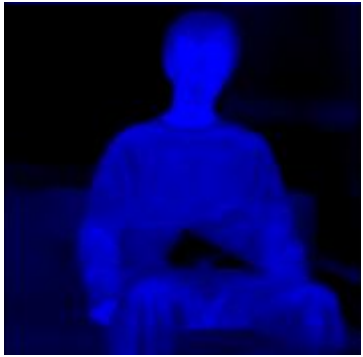
Figure 7 : HSV image

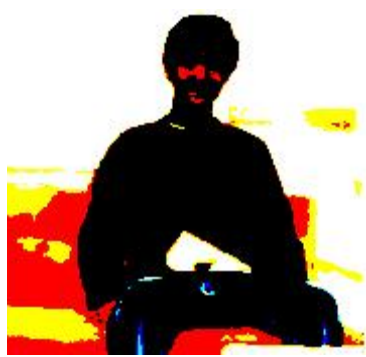
Figure 8 : Fused image

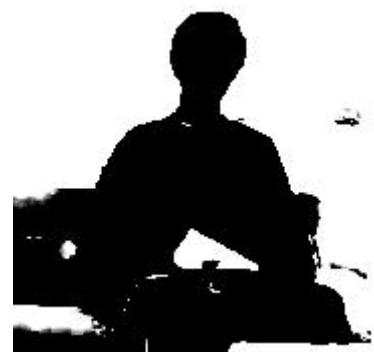
Figure 9: Fused gray image

This steps is required for the next step in which we use a binarization technique. There are several binarization techniques among them Otsu, Bernsen, savala , th-mean, niblack and iterative partitioning as a framework method are showing good result for this type of image. Here we use Otsu method which is a global Thresholding method i.e threshold value are calculated locally and get the result, no extra threshold value is added here. Extract this weapon portion by calculating all connected area component then remove too small component according to the area values. This only weapon portion binary image is shown in **figure 10**. Let us we want to show the weapon in the actual RGB visual image. The weapon binary images are stored into three different components because we want multiply it with three dimensional RGB image. Multiply individual element to element between two matrixes. In this step we detect weapon with visual RGB image is shown in **figure 11.**

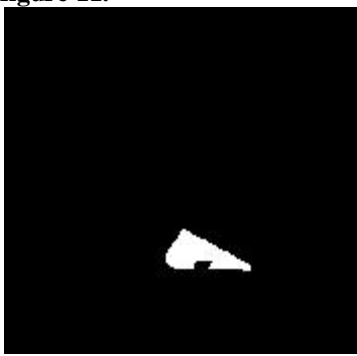
Figure 10 : Weapon in binary image

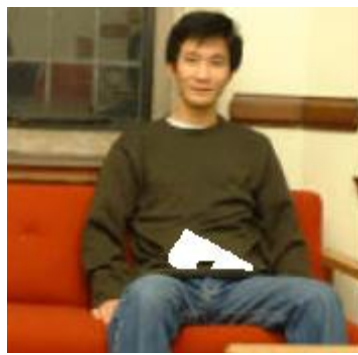
Figure 11 : Weapon in visual image

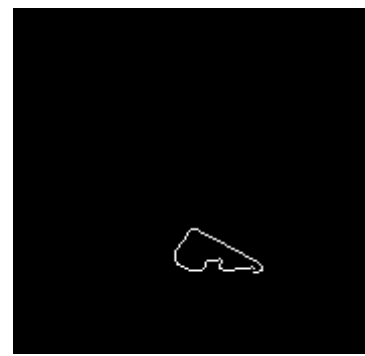
Figure 12 : Contour of the Weapon image

Contour detection is used to detect edges of weapon from the weapon binary image. Edge detection refers to the process of identifying and locating sharp discontinuities in an image. The discontinuities are abrupt changes in pixel intensity which characterize boundaries of objects in a scene. There is an extremely large number of edge detection operators available, each designed to be sensitive to certain types of edges. Here we use canny edge detection techniques. The Canny edge detection algorithm is known to many as the optimal edge detector. Canny's edge detection algorithm is computationally more expensive compared to Sobel, Prewitt and Robert's operator. However, the Canny's edge detection algorithm performs better than all these operators under almost all scenarios. This contour detection of concealed weapon is shown in **figure 12**.
Then this binarizes contour image are divided into three component and multiply as before and get contour with visual RGB image which is shown in **figure 13** where we can see the concealed weapon under person clothes easily.

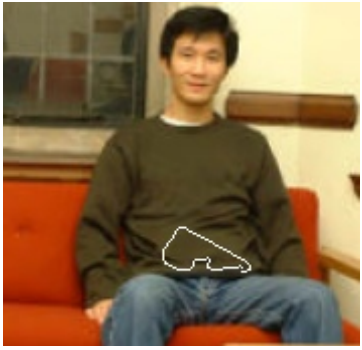

Figure 13 : Contour with Visual Image

Flowchart of the proposed methods is shown below:

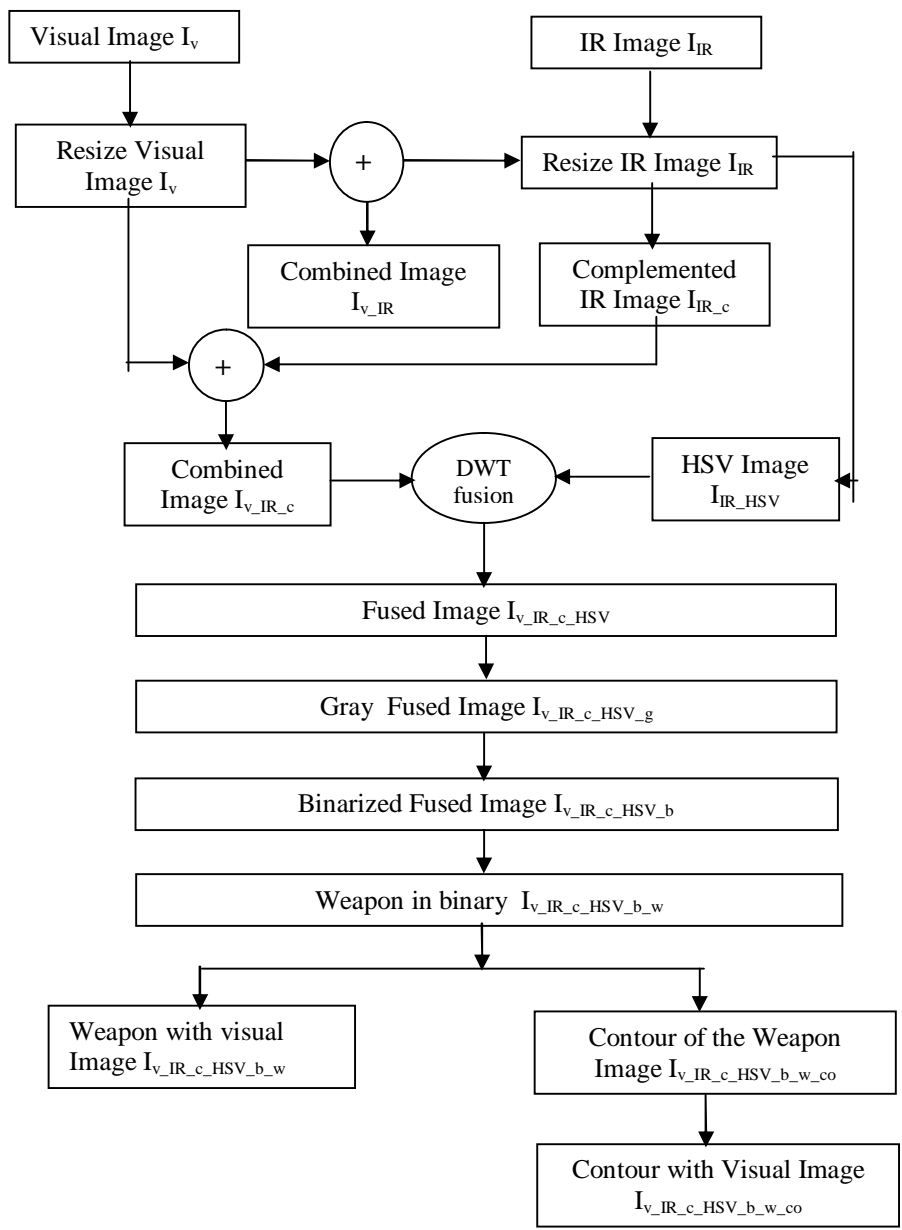

**Conclusion**

In this paper we introduce a color image fusion technique for CWD where we fuse a visual RGB image and IR image. We can able o detect the weapon concealed under person's clothes and bags. But infrared radiation can be used to show the image of a concealed weapon only when the clothing is tight, thin, and stationary. For normally loose clothing, the emitted infrared radiation will be spread over a larger clothing area, thus decreasing the ability to image a weapon. We try to solve this problem in our future work.

**References:**
1. Hua-Mei Chen, Seungsin Lee, Raghuveer M. Rao, Mohamed-Adel Slamani, and Pramod K. Varshney. : A tutorial overview of development in imaging sensors and processing. IEEE SIGNAL PROCESSING MAGAZINE, pp.52-61, MARCH 2005.
2. Yu-Wen Chang ; Michael Johnson. : Portable Concealed Weapon Detection Using Millimeter Wave FMCW Radar Imaging. Federal funds provided by the U.S. Department of Justice August 30, 2001.
3. Z. Xue, R. S. Blum, and Y. Li. : Fusion of Visual and IR Images for Concealed Weapon Detection1. U. S. Army Research Office under grant number DAAD19-00-1-0431, pp 1198-1205.
4. Sudipta Roy and Prof. Samir K. Bandyopadhyay. : Visual Image Based Hand Recognitions. Asian JournalOf Computer Science And Information Technolog(AJCSIT)y1:4 (2011), pp.106 – 110.
   http://innovativejournal.in/index.php/ajcsit/article/view/94
5. Mohamed-Adel Slamani , Pramod K. Varshney , David D. Ferris. : Survey of Image Processing Techniques Applied to the Enhancement and Detection of Weapons in MMW Data. SPIE Vol. 4719 (2002).
6. Zhiyun Xue, Rick S. Blum. : Concealed Weapon Detection Using Color Image Fusion. ISIF, pp-622-627,2003.
7. R. C. Gonzalez, R. E. Woods. : Digital Image Processing. Second Edition, Prentice Hall, New Jersey 2002.
8. Otsu, N.: A threshold selection method from gray-level histogram. IEEE Trans. Syst. Man Cybern. 9, 62–66 (1979)
9. Niblack,W.: An Introduction to Digital Image Processing. pp. 115–116. Prentice Hall, Eaglewood Cliffs (1986)
10. Sauvola, J., Pietikainen, M.: Adaptive document image binarization. Pattern Recogn. 33(2), 225–236 (2000)
11. Manjusha Deshmukh, Udhav Bhosale.: Image Fusion and Image Quality Assessment of Fused Images. International Journal of Image Processing (IJIP), pp. 484-508,Volume (4): Issue (5).
12. M. Aguilar, and J. R. New. : Fusion of multi-modality volumetric medical imagery. ISIF 2002, pp. 1206-1212.
13. Sudipta Roy, Prof. Samir K. Bandyopadhyay, "*Contour Detection of Human Knee*", International Journal of Computer Science Engineering and Technology (IJCSET) ,September 2011 , Vol 1, Issue 8,pp. 484-487.
14. Chu-Hui Lee and Zheng-Wei Zhou . : Comparison of Image Fusion based on DCT-STD and DWT-STD. Proceedings of the International Multiconference of Engineers and computer scientists 2012,vol I, IMECS 2012, Hong Kong.
15. Sudipta Roy, Prof. Samir K. Bandyopadhyay, "*Detection and Quantification of Brain Tumor from MRI of Brain and it's Symmetric Analysis*" ,International Journal of Information and Communication Technology Research(IJICTR), pp. 477-483,Volume 2, Number 6, June 2012.
16. 14    Stavri Nikolov_ Paul Hill_ David Bull_ Nishan CanagarajahWAVELETS FOR IMAGE FUSION Image    Communications Group Centre for Communications Research University of Bristol.